# Predicting Stress in Remote Learning via Advanced Deep Learning Technologies

Daben Kyle Liu
Belmont High School
Belmont, MA, USA
dabenkyle.liu@gmail.com

*Abstract*—COVID-19 has driven most schools to remote learning through online meeting softwares such as Zoom and Google Meet. Although this trend helps students continue learning without in-person classes, it removes a vital tool that teachers use to teach effectively: visual cues. By not being able to see a student's face clearly, the teacher may not notice when the student needs assistance, or when the student is not paying attention. In order to help remedy the teachers of this challenge, this project proposes a machine learning based approach that provides real-time student mental state monitoring and classifications for the teachers to better conduct remote teaching. Using publicly available electroencephalogram (EEG) data collections, this research explored four different classification techniques: the classic deep neural network, the traditionally popular support vector machine, the latest convolutional neural network, and the XGBoost model, which has gained popularity recently. This study defined three mental classes: an engaged learning mode, a confused learning mode, and a relaxed mode. The experimental results from this project showed that these selected classifiers have varying potentials in classifying EEG signals for mental states. While some of the selected classifiers only yield around 50% accuracy with some delay, the best ones can achieve 80% accurate classification in real-time. This could be very beneficial for teachers in need of help making remote teaching adjustments, and for many other potential applications where in-person interactions are not possible.

## I. INTRODUCTION

Most schools have been forced into some form of remote learning due to the COVID-19 pandemic. Video conferences have been helpful for teachers to safely keep in touch with the students, and have increasingly become a replacement for teaching in a traditional classroom. However, one of the big challenges in online teaching is the loss of the ability to visually observe students in real-time. By being in the same classroom, experienced teachers can acutely sense whether students are paying attention, and more importantly whether students are understanding the materials and concepts that are being taught. With those observations, teachers can adjust the delivery method or the teaching speed on the fly and create an effective teaching environment. With video conferencing softwares such as Zoom and Google Meet, it is harder to observe students and generally more difficult for the teacher to judge how effective the teaching materials are being delivered. One of the big problems is that the teacher cannot easily see every student from the small squares on a small computer screen. Even then, the student can mute themselves and stop their video, which prevents the teacher from picking up any signs of discomfort from the student. On top of that, students who need help may not want to speak up due to the unfamiliarity of speaking in a remote setting. In an attempt to resolve this challenge, schools sometimes send out surveys to students to identify the difficulty of the homework or the learning curriculum in general. However, the survey results could be quite biased in that the students may not be willing to answer truthfully. Combined with the fact that these surveys are time-consuming for the teacher and generally not timely for the students, the challenge remains for teachers to effectively conduct remote teaching. Teachers are desperately in need of new tools to help capture a student's mental state with the introduction of remote teaching.

Traditionally it is accepted that one of the most reliable ways of determining the mental state is to examine the reflection of various measurements on brain activity, which can be represented by Electroencephalogram (EEG) signals. EEG has frequently been used in clinical diagnoses, biomedical research, and behavior analyses[1-3]. Stress, anxiety, and pleasure are a few of the examples where EEG signals have been proven to show a high correlation.

The recent rapid development and adoption of deep neural networks in the machine learning field has opened a vast number of opportunities in highly accurate data analysis, data mining, and data classification. This movement inevitably penetrated the area of Biomedical Science too, especially in EEG data analysis. Deep Neural Networks (DNN) have been used to detect stress levels in construction workers using EEG signals as an input [4]. The researchers in this research used a convenient wearable EEG headset to obtain the EEG recording, then measured levels of cortisol to get a stress reference. After that, they passed the raw EEG through a preprocessor, and finally, put it through a Convolutional Neural Network (CNN) to determine the stress levels. They achieved about a 64% accuracy with a CNN, and 87% with a DNN. Some scientists proposed a method that uses EEG signals to detect stress and introduces stress reduction techniques by adding interventions into their method [5]. They used the support vector machine (SVM) as one of their classification techniques. Others proposed an EEG-based stress recognition framework that considers each subject's brainwave patterns and continuously updates itself based on the new input signals in near real-time[6]. The framework first removes EEG signal artifacts, then extracts a broad range of EEG signals, and finally applies different online multitask learning (OMTL) algorithms to recognize the individuals' stress in near real-time. Most of the work presented 70-87% accuracy achieved using various machine learning methods. However, none have specifically studied EEG signal usage in a student learning scenario.

In this research, a study was performed where a student's mental states in a learning environment were classified into 3 distinctive states:

- Engaged learning: where students can grasp the taught concept and follow the teacher's teachings without difficulty
- Confused learning: where students are not able to understand the material being taught, even though they are still engaged in learning
- Relaxed state: where students are disengaged with the current learning (not paying attention)

With EEG signals used as the input, 4 different types of popular classifiers were explored in order to find the fastest model that can output accurate results: SVM, DNN, CNN, and XGBoost. The first 3 classifiers were chosen due to their popularity in literature related to EEG signal classification. The 4th one, XGBoost, is a decision tree style machine learning model that can optimize arbitrary loss functions with gradient boosting. It has recently gained lots of popularity due to surprisingly high accuracy readings achieved in various data science competitions. While the literature research did not yield any mentions of using XGBoost in EEG signal processing, the approach seems suitable nonetheless for the task.

In the remainder of this paper, I will first describe the data sources used in this study in detail. Then I will briefly introduce the different classifiers and how the input features are constructed. Experimental results will be presented next. Finally I will conclude with some discussions about future work.

## II. Data

The search for data to be used in this project focused on publicly available EEG datasets. However, there were not many available that were related to students. One of the most relevant datasets was found on Kaggle[9], which recorded EEG signals when students were presented with teaching videos of designated difficulty levels. Another dataset relevant to this study was from OpenNeuro[16], which recorded EEG signals during meditation. The detailed descriptions of both datasets are provided here, as well as the preprocessing steps needed to reconcile the different methods of both datasets.

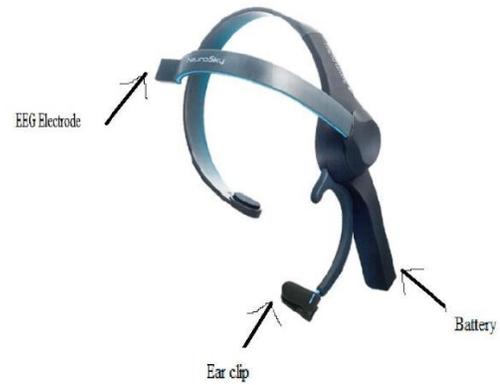

*Figure 1. Mindset Headset. Only Has the Fp1 Channel [17]*

### A. Kaggle Student EEG Data

In [9], the authors collected EEG signal data from 10 college students (subjects) while they watched 10 video clips. These videos are one minute long each and focus on topics varying from basic Algebra to Stem Cell Research. These topics are either within the education level of the subject, or at a much more advanced level than the subject's current grade and designed to confuse the subject. The EEG signal was recorded by a single-channel wireless MindSet, shown in Figure 1, that measured activity over the frontal lobe. The MindSet measures the voltage between an electrode resting on the forehead and two electrodes (one ground and one reference) each in contact with an ear. Previous research has shown that this type of wireless EEG device can produce compatible quality signals as traditional wired EEG system[15]. The EEG bands are compiled every 0.5 seconds. After each session, the student rated his/her confusion level on a scale of 1-7, where one corresponded to the least confusing and seven corresponded to the most confusing. These labels are further classified into binary labels of whether the students are confused or not.

One limitation of this data set is that there is no raw data, and only the limited power spectrum of the EEG frequency bands are available. As described by the author, the frequency band readings were directly from the device and no raw data was recorded. Besides the usual 5 bands - Delta, Theta, Alpha, Beta, Gamma, the dataset further divides the Alpha, Beta, and Gamma bands into 2 equal frequency bands each, forming Alpha1, Alpha2, Beta1, Beta2, Gamma1, and Gamma2 bands. As a result, the same band division was used for other data sets so that they are compatible as input features to the same ML classifiers.

## B. OpenNeuro EEG Data

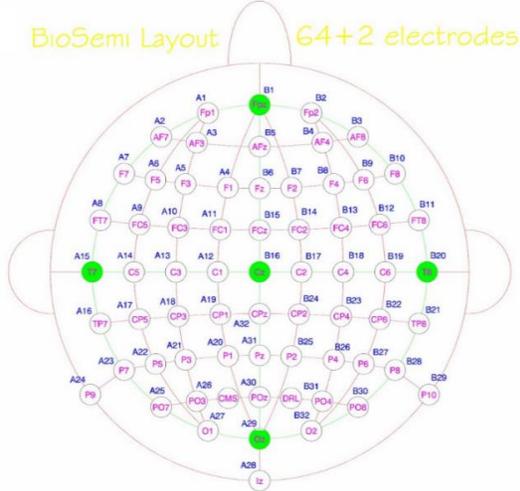

*Figure 2. Biosemi Headcap, 64 channels, 10/20 layout* [18]

In [10], expert and non-expert meditators alike were asked to meditate continuously throughout the experiment in a seated meditation position. Once all subjects were comfortably seated in their meditative posture, they were instructed to begin their meditation. All practitioners began with an initial body scan as they relaxed into their seated posture. Experience-sampling probes were presented at random intervals ranging from 30 to 90 seconds throughout the duration of the experiment. The time range of the experiment lasted from 45 min to 1 hour and 30 minutes, as some subjects were willing and able to sit comfortably for longer periods of time. The authors collected data using a 64-channel Biosemi system and a Biosemi 10–20 head cap, as shown in Figure 2, montage at 2048 Hz sampling rate. All electrodes were kept within an offset of 15 using the Biosemi ActiView data acquisition system for measuring impedance. Respiration, heart rate (ECG/HRV) and galvanic skin response (GSR) were also recorded, but results from these data were not reported in the dataset. Raw data is available for this data set.

It was noted in the dataset documentation that not all subjects are experts in meditation. Given that the people labeled non-experts may produce some unreliable data, in this experiment, only the expert labeled data were used to remove any noise.

## C. Data Preprocessing

Besides the general high-pass filtering to remove DC noise, extra preprocessing was needed to match the two datasets. These are the 3 required steps:

1. Channel compatibility. Because the Kaggle dataset only measures the Fp1 channel of the subjects only the A1 channel recording from the OpenNeuro dataset was used in the study, which recorded EEG signals from the same area as the Fp1 channel (see Figure 2).
2. Feature extraction: For OpenNeuro data processing, 8 bands of power spectrum were computed, rather than the usual 5 bands in order to match the Kaggle dataset's higher resolution for Alpha, Beta, and Gamma bands. In addition, this project used 2 dimensions for gender and 1 dimension for age, which leads to 11-dimensional features used in my experiments.
3. Sampling rate: An analysis using FFT every 0.5 seconds on the OpenNeuro raw data were performed, resulting in features with the same sampling rate as the Kaggle dataset.

## III. CLASSIFIERS

In this work, 4 different types of popular classifiers were studied: SVM, fully connected DNN, CNN, and XGBoost. Publicly available Python packages were used in this study, which include TensorFlow, SKLearn, and XGBoost. In this section, a brief description of each classifier and their strengths are provided.

### A. SVM

In machine-learning, support-vector machines are a set of supervised learning models with maxium margin objectives. In the simplest case of linearly separable set of training data $\{(\bar{x}_1, y_1), (\bar{x}_2, y_2), \ldots, (\bar{x}_n, y_n)\}$, where $\bar{x}_i$ is the input vector, and $y_i$ is the class label of 1 or -1, the hyperplane of label 1 and -1 can be described as:

$W^T \bar{x} - b = 1$  hyperplane for label 1

$W^T \bar{x} - b = -1$  hyperplane for label -1

where $W$ is the normal vector of the hyperplane and $b$ is the constant bias. The distance of the two hyperplane is thus $\frac{2}{\|W\|}$. To find the maximum margin (distance) between the two hyperplanes would mean to minimize $\|W\|$ [11]. The formular can be extended to more complex nonlinear classifications with more than 2 classes.

SVM was very popular in the 80s and 90s because it provides high accuracy with relatively less required computation. It is also widely used in biomedical fields for EEG signal classification [4][5][6]. It would be very interesting to see how it works with this project's target of mental states during learning.

### B. DNN

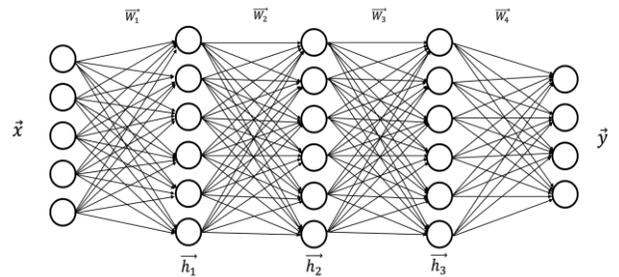

*Figure 3. A Fully Connected (Dense) DNN*

Deep Neural Network is sometimes interchangeably used with "Deep Learning". It has made possible for the last 10 years due to its phenomenal success in many different fields of machine learning, significantly improving the state-of-the-art accuracy.

As shown in Figure 3, neural network is a layered network of neurons connected by weighted links $\overrightarrow{W_l}$. In the picture, each circle is an artificial neuron. There is a 5-dimentional input layer ($\vec{x}$), 3 hidden layers denoted by $\vec{h_l}$ with 6 neurons at each layer, and finally 4 output nodes indicating a 4-class label ($\vec{y}$) problem. Each neuron takes the weighted summation of input from all neurons of previous layer and pass the value through a non-linear activation function which provides neural network its learning power. "Deep" in DNN just means a lot of hidden layers and hidden nodes.

In this paper, the term DNN is used to refer to a generic fully connected Deep Neural Network. Also, due to the limited amount of data, best results were obtained with 2 hidden layers, each layer consisting of 450 nodes. The size of the resulting DNN is not excessively deep, even though this size would have been considered computationally prohibitive 10 years ago.

*C. CNN*

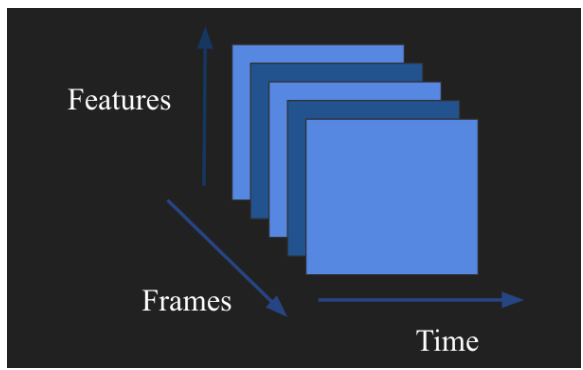

*Figure 4. CNN Feature Map*

In deep learning, a convolutional neural network (CNN) is a class of deep neural networks, usually applied to analyzing visual imagery. They are also known as shift invariant or space invariant artificial neural networks (SIANN), based on their shared-weights architecture and translation invariance characteristics. They have applications in image and video recognition, recommender systems, image classification, medical image analysis, natural language processing, and financial time series [13]. CNNs are probably one of the most popular classification techniques due to the high accuracy they generally achieve. By design, CNNs are also regularized so that the training would unlikely be overfitting.

The input to CNNs is different from the other in that it requires an input of a 4-tuple: [number of images, image width, image height, number of channels]. To accommodate the requirement, I applied a sliding window of 20 frames (10 seconds) which generates a 11x20 feature map for input to CNNs, as well as sampling the feature every 11 frames.

*D. XGBoost*

XGBoost stands for eXtreme Gradient Boosting [14]. Gradient boosting is a machine learning technique for regression and classification problems, which produces a prediction model in the form of an ensemble of weak prediction models, typically decision trees. It builds the model in a stage-wise fashion like other boosting methods do, and it generalizes them by allowing optimization of an arbitrary differentiable loss function [8].

XGBoost is particularly popular in Kaggle competitions and has won several of them. It is computationally efficient while yielding particularly good results on structured and tabular data. Even though no application or literature mentioned XGBoost as one of the selected tools for EEG signals, I would like to give it a try for my experiment.

*E. Model Setup*

This section will describe the model parameters and topology. I have experimented with a set of different configurations for each classifier. What I provide here has yielded the best accuracy.

- **SVM:** To set up the SVM, I used scikit's sklearn.svm.SVC. I changed the regularization parameter to 0.5 instead of the default 1.0. Everything else was left as is.
- **DNN:** I used Tensorflow's pre-made tf.estimator.DNNClassifier, with 2 hidden layers of 450 nodes. The feature columns and number of classes were changed to fit the training and testing sets.
- **CNN:** I used TensorFlow's constructor tf.models.Sequential as the framework to construct my CNN. The convolutional part of the model contains two Conv2D layers, with a MaxPooling2D layer in between them. Both Conv2D layers have kernel sizes of 3*3 and Relu activation functions. The first Conv2D layer has 32 filters and an input shape of (20,11,1). The second Conv2D layer has 64 filters. The MaxPooling2D layer has a pooling size of 2*2. After the convolutional part comes a flattening layer, followed by two dense layers. The first dense layer contains 32 nodes, with a Relu activation function. The second and last dense layer contains 3 output nodes, one for each class.
- **XGBoost:** I did not change any of the default settings that came with the XGBoost framework.

IV. EXPERIMENTAL RESULTS

After compiling all the collected data into a single data frame, the data were shuffled and split into two sets, where 80% were used for training, and 20% were used for testing. The split resulted in a test set size of 1451 samples. Then each classifier was trained and tested. The accuracy results are shown in Figure 5. Note that a random classification with a random number generator was also performed for reference. A valid machine-learning classifier would result in an accuracy higher than the random accuracy.

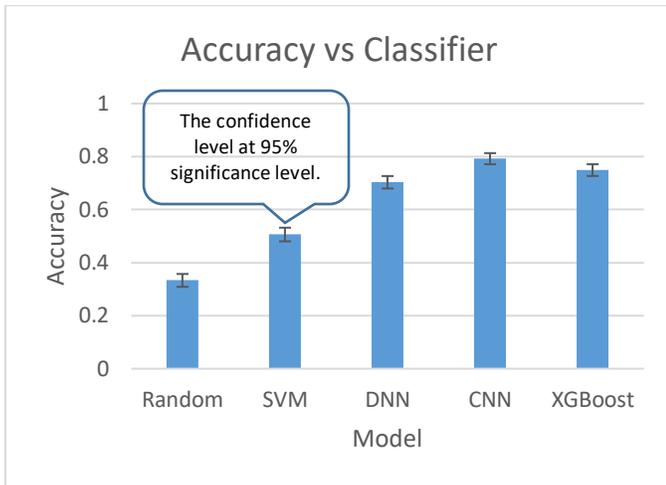

Figure 5. Accuracy Ratings for Each Model, Random Guess at 33%

As shown in the accuracy graph, all classifiers performed better than the random classifier, indicating they all learnt from the data as expected. The CNN has the best accuracy. At 79.2%, CNN does a decent job in classifying the mental states, and can possibly be used in a real-world scenario. The model that did the worst was the SVM, with a low score of 50.6%. However this is understandable, as SVM is efficient for binary classification, and may have trouble distinguishing more than two classes. The most surprising result is the accuracy of the XGBoost model. Even though it wasn't mentioned in any existing EEG classification work, it got an accuracy of 74.9%, which is better than the classic Deep Neural Network's accuracy of 70.3%.

The confidence intervals were also calculated and displayed in Figure 5, where Gaussian Distribution is assumed, and 95% significance level is used:

$$CI = 1.96 * \sqrt{\frac{accuracy * (1 - accuracy)}{n}}$$

Equation 1. Confidence Interval for Classification at 95% level

where n is the sample size for the test set, which equals 1451 in this study. The resulting confidence intervals are between 2.0% to 2.5% with an average error margin of ±2.31%.

While the accuracy ratings show promise, it is necessary to consider the speed of each classifier if they are considered for practical applications. In order to do so, I measured the time it took for each model to evaluate the same test set. To measure the trade-off between accuracy and speed, I defined a metric, "Potential", which is denoted as follows (Equation 2):

$$Potential = accuracy/time$$

Equation 2. Accuracy is in Decimals, Time is in Seconds

The higher the accuracy, and the shorter the testing time, the higher Potential the model would have. The Potential of the random classifier is 0 by definition since it has 0 potential to be used in practice.

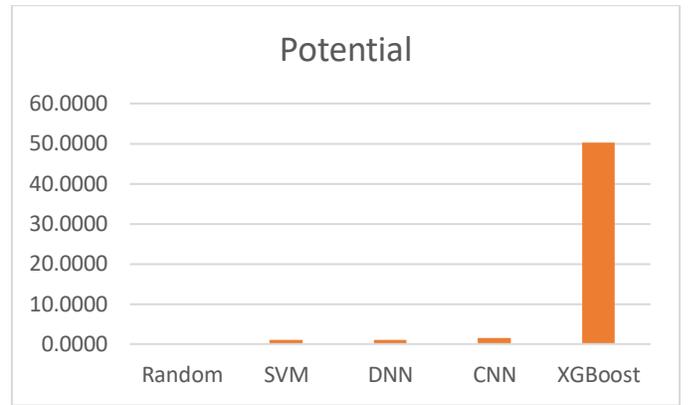

Figure 6. Potential of Each Model, Random=0.33/0

Figure 6 shows the Potential of the 4 classifiers. While most of the models took around half a second to evaluate the testing data, the XGBoost model was able to do it in 0.015 seconds. Along with the high accuracy ratings rivaling that of the CNN, XGBoost has the best potential by far for classifying EEG signals in real time. I believe the potential of XGBoost for EEG signal processing should be further explored.

V. CONCLUSION

COVID-19 has given teachers the added challenge of not being able to clearly discern students in need of help. To solve this problem, I propose machine-learning based automatic mental state classification and monitoring as a new method assisting teachers in a remote learning environment. Four different Machine Learning techniques were explored in this study to classify learning and relaxing mental with EEG signals. From the accuracy alone, it was clear that the CNN provided the best accuracy. However, the XGBoost model was able to create accuracies close to the CNN in a much smaller time frame. Given that there are scarcely any prior work using XGBoost for EEG data, I hope this study provides a new promising direction. At close to 80%, I think both the CNN and XGBoost model are usable for detecting students' mental states during remote learning and can provide useful information to teachers and students alike.

VI. FUTURE WORK

Given the limitation of the data sources available to me, I was confined to a small set of features. In the future, with more targeted EEG data collections in learning environments that provide raw recordings, the feature set can be greatly enriched, which would hopefully result in much higher accuracy than 79.2%.

Imagine a real-world remote learning setting where each student is wearing a noise-cancelling headset with a single probe on the frontal lobe area for EEG collection. EEG signals are sent to the servers as part of the video/audio signal. Mental classification is conducted on the servers and sent to both the student and the teacher. A confused student would be highlighted at the teacher(host)'s screen so that the teacher can take some measures to help the student. On the other hand, a very relaxed student would result in a warning sign on his/her screen. If this can become a reality, then remote learning will become even more viable, during and after the COVID-19 pandemic.

This study aims for the scientific research and practical viability of the approach in technical terms. However, I recognize that EEG signals and mental states are private to the person and must be protected if and when the application is implemented in reality.